%% file: main.tex
\documentclass[10pt,twocolumn,letterpaper]{article}

\usepackage[pagenumbers]{cvpr} % To force page numbers, e.g. for an arXiv version


\definecolor{cvprblue}{rgb}{0.21,0.49,0.74}

\usepackage[colorlinks,
            linkcolor=red,
            anchorcolor=blue,
            citecolor=green
            ]{hyperref}

\def\paperID{3912} % *** Enter the Paper ID here
\def\confName{CVPR}
\def\confYear{2025}

\usepackage[ruled,linesnumbered] {algorithm2e}
\usepackage{multicol}
\usepackage{multirow}
\usepackage{booktabs}
\usepackage{colortbl}
\usepackage{pifont}
\definecolor{coldgrey}{RGB}{128,128,105}
\usepackage{stfloats}

\title{A Simple Data Augmentation for Feature Distribution \\Skewed Federated Learning}

\author{Yunlu Yan$^1$, Huazhu Fu$^2$, Yuexiang Li$^3$, Jinheng Xie$^4$, Jun Ma$^{1,6}$, Guang Yang$^5$, Lei Zhu$^{1,6}$ \\
{\small$^1$The Hong Kong University of Science and Technology (Guangzhou) \quad $^2$IHPC, A*STAR}  \\
{\small$^3$Guangxi Medical University \quad  $^4$National University of Singapore \quad $^5$Imperial College London}\\
{\small$^6$The Hong Kong University of Science and Technology}
}

\begin{document}
\maketitle
\begin{abstract}
Federated Learning (FL) facilitates collaborative learning among multiple clients in a distributed manner and ensures the security of privacy. However, its performance inevitably degrades with non-Independent and Identically Distributed (non-IID) data. In this paper, we focus on the feature distribution skewed FL scenario, a common non-IID situation in real-world applications where data from different clients exhibit varying underlying distributions.  This variation leads to feature shift, which is a key issue of this scenario.  While previous works have made notable progress, few  pay attention to the data itself, \textit{i.e.}, the root of this issue. The primary goal of this paper is to mitigate feature shift from the  perspective of data. To this end, we propose a simple yet remarkably effective input-level data augmentation method, namely FedRDN, which randomly injects the statistical information of the local distribution from the entire federation into the client's data. This is beneficial to improve the generalization of local feature representations, thereby mitigating feature shift. Moreover, our FedRDN is a plug-and-play component, which can be seamlessly integrated into the data augmentation flow with only a few lines of code. Extensive experiments on several datasets show that the performance of various representative FL methods can be further improved by integrating our FedRDN, demonstrating its effectiveness, strong compatibility and generalizability. Code will be released.
\end{abstract}

\section{Introduction}
% Background of FL
Federated Learning~\cite{li2020survey,li2021survey,liu2024vertical} (FL) has been a de facto solution for distributed learning, and attracted wide attention from various communities~\cite{pengfederated,lee2021preservation,fang2022robust,dong2022federated}. It utilizes a credible server to communicate the privacy irrelevant information, \textit{e.g.}, model parameters, thereby collaboratively training the model on multiple distributed data clients, while the local data of each client is invisible to others. Such a design is simple yet can achieve superior performance and preserve privacy. However, it is inevitable to suffer non-Independent and Identically Distributed (non-IID) data when we deploy FL in real-world applications, which will greatly degrade the performance of the trained model~\cite{karimireddy2020scaffold,li2020federated,luo2021no}.

% pose the problem
Since non-IID data is widely common in many real-world cases, many researchers try to address this issue and improve the practicability of FL~\cite{karimireddy2020scaffold,li2020federated,hsu2019measuring,reisizadeh2020robust}. However, most of them try to design a general FL method that can handle different non-IID scenarios. In fact, such a solution is suboptimal, due to the discrepancy between different non-IID scenarios, such as feature distribution skew and label distribution skew~\cite{li2022federated}. Hence, recent studies try to design special FL algorithms for different non-IID scenarios~\cite{zhang2022federated,zhou2023fedfa}. In this work, we focus on solving feature distribution skew (Definition in \S\ref{sec:pre}), a typical non-IID scenario in FL, as the data of discrete clients are always collected from different devices or environments, incurring different underlying distributions. For instance, different hospitals possess MR images scanned by different devices and different phones store images of different styles (such as cartoon images and natural images). Such data situation leads to inconsistent feature distribution across different clients, namely \textbf{\textit{feature shift}}~\cite{li2021fedbn,zhou2023fedfa}, resulting in serious performance degradation. To address this problem, FedBN~\cite{li2021fedbn} learns the specific batch normalization layer parameters for each client. HarmoFL~\cite{jiang2022harmofl} utilizes the potential knowledge of medical images in the frequency domain to mitigate feature shift.

% motivation
While feature distribution skew has been explored in FL, little attention has been paid to the data itself. When it comes to data processing, the common practice~\cite{jiang2022harmofl,zhou2023fedfa,li2021fedbn} is to integrate some traditional data operations in the data augmentation flow (\textit{e.g.}, \textit{transforms.Compose}() in Pytorch) of each client like flipping, cropping and rotating. This overlooks the effectiveness of data augmentation for FL at the input level, \textit{i.e.}, data augmentation flow, which leaves an alternative space to explore. Since the root cause of feature shift lies in the divergence of local data distributions, we ask two questions: 1) \textbf{\textit{Why not proactively resolve it by directly processing the data?}}
2) \textbf{\textit{Can we design an FL-specific data augmentation technology that can be integrated into the data augmentation flow?}} To answer these questions, we try to design a plug-and-play input-level data augmentation technique.

Although it may appear straightforward, effectively implementing data augmentation in FL poses a significant challenge, due to the lack of direct access to the external data of other clients. Therefore, how to inject global information into augmented samples and thereby mitigate the distribution bias between different local datasets is the core of the challenge.
% The main challenge lies in how to inject global information into augmented samples, thereby mitigating the distribution bias between different local datasets. 
In this regard, FedMix~\cite{yoon2021fedmix} extends Mixup to FL for label distribution skew by incorporating the mixing of averaged data across multiple clients. However, it is only suitable for the classification task and its effectiveness for feature distribution skew has not been demonstrated. Furthermore, permitting the exchange of averaged data introduces privacy concerns. In a more recent study, a data augmentation approach, namely FedFA~\cite{zhou2023fedfa}, is proposed for feature distribution skew, which mitigates feature shift through feature augmentation based on the statistics of latent features. Even though, it requires modifications to the network structure, which cannot be seamlessly integrated into the data augmentation flow.  Besides, FedFA adds additional computational and communication overhead, which can be constrained in resource-limited scenarios.

In this work, we propose a novel federated data augmentation technique, called FedRDN. We argue that the local bias is rooted in the model trained on the limited and skewed distribution. In centralized learning, we can collect data from many different distributions to learn generalized feature representations. This motivates us to augment the data to a more abundant distribution, which indirectly reduces the difference between different local distributions. To achieve that in the setting of FL, our FedRDN extends the standard data normalization to FL by randomly injecting statistics of local datasets into augmented samples, which is based on the insight that statistics of data are the essential characteristic of data distribution. It enables clients to access a wider range distribution of data, thereby enhancing the generalizability of local feature representations. Our FedRDN is a remarkably simple yet surprisingly effective method. It is a non-parametric approach that incurs negligible computation and communication overhead, without requiring modifications to the network structure, and seamlessly integrates into the data augmentation pipeline with just a few lines of code. After employing our method, significant improvements have been observed across various typical FL methods. In a nutshell, our contributions are summarized as follows:

\begin{itemize}
	\setlength{\itemsep}{5pt}
	\setlength{\parsep}{-2pt}
	\setlength{\parskip}{-0pt}
	\setlength{\leftmargin}{-15pt}
 
\item We explore the input-level data augmentation technique for feature distribution skew, which gives more insights into how to understand and solve this problem.
 
\item We propose a novel plug-and-play data augmentation technique, FedRDN, which can be easily integrated into the data augmentation flow to mitigate feature shift for feature distribution skewed FL.
\item We conduct extensive experiments on two classification datasets and an MRI segmentation dataset to demonstrate the effectiveness and generalization of our method, \textit{i.e.}, it outperforms traditional data augmentation techniques and improves the performance of various typical FL methods.

\end{itemize}
\section{Related Work}
\noindent{\textbf{Federated Learning with Non-IID Data.}} FL~\cite{zhang2022robust,zhang2022game,li2021survey} allows multiple discrete clients to collaborate in training a global model while preserving privacy. The pioneering work, FedAvg~\cite{mcmahan2017communication}, is the most widely used FL algorithm, yielding success in various applications. However, the performance of FedAvg is inevitably degraded when suffering non-IID data~\cite{li2020federated,karimireddy2020scaffold}, posing a fundamental challenge in this field. To address this, a variety of novel FL frameworks have been proposed, with representative examples such as FedProx~\cite{li2020federated}, Scaffold~\cite{karimireddy2020scaffold}, and FedNova~\cite{wang2020tackling}. They improved FedAvg by modifying the local training~\cite{li2021model,acarfederated} or model aggregation~\cite{yurochkin2019bayesian,lin2020ensemble} to increase stability in heterogeneous data environments. Despite the progress, most of them ignore the differences in various non-IID scenarios, like the heterogeneity of label distribution skew and feature distribution skew lies in label and images~\cite{li2020survey}, respectively. Therefore, recent studies attempt to more targeted methods for different non-IID scenarios. For example, FedLC~\cite{zhang2022federated} addressed label distribution skew by logits calibration. As for feature distribution skew, the target of this work, FedBN~\cite{li2021fedbn} learned the heterogeneous distribution of each client by personalized batch normalization layer parameters. In addition, HarmoFL~\cite{jiang2022harmofl} investigated specialized knowledge of frequency to mitigate feature shift. More recently, FedPCL~\cite{tan2022federated} employed a pre-trained model to reduce the number of learnable parameters and applied prototype-wise contrastive learning to regularize feature representation learning across different clients. This offers an effective solution for training large models in the feature distribution skewed FL scenario. Motivated by prototype learning, FPL~\cite{huang2023rethinking} utilized clustering to acquire unbiased class prototypes and then alleviated feature shift through prototype and local embedding alignment. In contrast to the aforementioned methods, ADCOL~\cite{pmlr-v202-li23j} introduced a novel adversarial collaborative learning approach to mitigate feature shift by replacing the model-averaging scheme with adversarial learning. Although achieving significant progress, they only focus on mitigating feature shift at the local optimization or global aggregation stage, neglecting the data itself which is the root of feature shift. Different from them, we aim to address this issue from the perspective of data.

\noindent\textbf{{Data Augmentation}} Data augmentation~\cite{kukavcka2017regularization,kumar2023advanced} is a widely used technique in machine learning, which can alleviate overfitting and improve the generalization of the model. For computer vision tasks, the neural networks are typically trained with a data augmentation flow containing various techniques like random flipping, cropping, and data normalization. Different from these early label-preserving techniques~\cite{zhong2020random}, label-perturbing techniques are recently popular such as MIXUP~\cite{zhangmixup} and CUTMIX~\cite{yun2019cutmix}. It augments samples by fusing not only two different images but also their labels. Except for the above input-level augmentation techniques, some feature-level augmentation techniques~\cite{li2021feature,li2021simple,Venkataramanan_2022_CVPR} that make augmentation in feature space have achieved success. Recently, some studies have started to introduce data augmentation techniques into FL. For example, FedMix~\cite{yoon2021fedmix} proposed a variant of MIXUP in FL, which shows its effectiveness in the label distribution skewed FL scenario. However, its effectiveness in the feature distribution skewed FL scenario has not been demonstrated. More importantly, FedMix requires sharing averaged local data which increases the risk of privacy. For feature distribution skew, FedFA~\cite{zhou2023fedfa} augmented the features of clients by the statistics of features. However, it operates on the feature level, which requires modifications to the network structure and incurs additional computation and communication cost. Different from FedFA, our proposed augmentation method operates on the input level. By this, our method can be a plug-and-play component, combined with different FL methods to further improve their performance, which shows stronger compatibility and flexibility. In addition, our method can further improve the performance of FedFA due to operating in different spaces.

\section{Methodology} \label{sec:method}
\subsection{Preliminaries} \label{sec:pre}
\noindent{\textbf{Federated Learning.}} Supposed that a FL system is composed of
 $K$ clients $\{C_1, C_2, \ldots, C_K\}$ and a central server. For client $C_k$ ($k\in [K]$), there are $n_k$ supervised training samples $\{x_i,y_i\}_{i=1}^{n_k}$, where image $x_i$ and label $y_i$ from a joint distribution $(x_i, y_i) \sim P_k(x, y)$. Besides, each client trains a local model $f(\boldsymbol{w}_k)$ only on its private dataset, where $\boldsymbol{w}_k$ is the parameters of the local model. The goal of FL is to learn a global model by minimizing the summation empirical risk of each client:
\begin{equation}
\min \mathcal{L} = \sum_{k=1}^{K} \gamma_{k} \mathcal{L}_{k}
\label{eq:1}, \quad \text{where} \quad \gamma_{k}= \frac{n_{k}}{\sum_{i=1}^{K}n_{i}}.
\end{equation}
To achieve this goal, at each communication round $t \in [T]$, the standard FL method, FedAvg~\cite{mcmahan2017communication}, performs $E$ epochs local training and the local objective $\mathcal{L}_k$ can be written as:
\begin{equation}
     \mathcal{L}_{k} = \frac{1}{n_k}\sum_{(x_i, y_i)\sim P_k} \ell(f(x_i;\boldsymbol{w}_k^t);y_i), 
     \label{eq:eq2}
\end{equation}
where $\ell$ is the loss function. After local training, it then averages the parameters of all local models to update the parameters of the global model, which can be described as:
\begin{equation}
\boldsymbol{w}_{G}^{t+1} =  \sum_{k=1}^{K} \gamma_{k} \boldsymbol{w}_{k}^{t}
\label{eq:2}.
\end{equation}
The updated parameters of global model $\boldsymbol{w}_{G}^{t+1}$ will be returned to each client as the initialization for the next round of training.

\noindent{\textbf{Feature Distribution Skew.}}  The underlying data distribution $P_k(x,y)$ can be rewritten as $P_k(y|x)P_k(x)$, and $P_k(x)$ varies across clients while  $P_k(y|x)$ is consistent for all clients. Moreover, different underlying data distributions lead to inconsistent feature distribution across different clients, thereby degrading the performance of the global model. 

\noindent{\textbf{Data Normalization.}} Normalization is a popular data preprocess operation, which can transform the data distribution to standard normal distribution. In detail, given an $C$-channel image $x \in \mathbb{R}^{C \times H \times W}$ with spatial size $(H \times W)$, it transforms image as:
\begin{equation}
 \hat{x} = \frac{x - \mu}{\sigma}, \quad \mu, \sigma \in \mathbb{R}^{C} 
\label{eq:3},
\end{equation}
where $\mu$ and $\sigma$ are channel-wise means and standard deviation, respectively, and they are usually manually set in experiential or statistical values from the real dataset.

% \subsection{Motivation}
% Since FL aggregates the learned knowledge from discrete clients, 

% it is straightforward to see that each client will bias to its skewed distribution.

\subsection{Federated Random Data Normalization}
In this section, we present the details of the proposed \textbf{F}ederated \textbf{R}andom \textbf{D}ata \textbf{N}ormalization (FedRDN) method. Different from previous FL methods, FedRDN focuses on mitigating the distribution discrepancy through the input-level data augmentation during the data processing stage. The goal of FedRDN is to let each client learn as many distributions as possible instead of self-biased distribution, which is beneficial to feature generalization. To achieve this, it performs explicit data augmentation by manipulating multiple-clients channel-wise data statistics during training at each client.  We will introduce the
detail of our method in the following. 

\begin{algorithm*}[t]
\caption{FedRDN}
\label{alg:fedrdn}
\KwIn{Number of Clients $K$, communication rounds $T$, epochs $E$}
\begin{multicols}{2}
\textbf{Compute  Statistic:} \\
 \For{client $k=1,2,...,K$ \textbf{parallelly}}{
    
        \For{$x_{i}^{k} \sim P_k$}{
            $\mu^k_{i} = \frac{1}{HW}\sum_{h=1}^{H}\sum_{w=1}^{W} x_{i}^{k,(h,w)}$ \\
            $\sigma^k_{i} = \sqrt{\frac{1}{HW}\sum_{h=1}^{H}\sum_{w=1}^{W} (x_{i}^{k,(h,w)}-\mu^k_{i})^2}$
        }
        $\mu^k = \sum_{i=1}^{n_k}\mu^k_{i} , \quad
         \sigma^{k} = \sum_{i=1}^{n_k}\sigma^k_{i}$ \\
    }
    \textbf{Return} $\{(\mu^k, \sigma^{k})\}_{k=1}^{K}$ \\
\newpage

\textbf{Data Augmentation:} \\
\For{round $t=1,2,...,T$}{
\For{epoch $e=1,2,...,E$}{
  \For{$(x_{i}^{k}, y_{i}^{k}) \sim P_k$}{
    $(\mu^j, \sigma^j) \sim \{(\mu^k, \sigma^{k}) \}_{k=1}^{K}$\\
    $\hat{x}^k_{i} = \frac{x_i^{k} - \mu^{j}}{\sigma^j}$ \\
     \tcp{training}
}}}
\end{multicols}
\end{algorithm*}

\noindent{\textbf{Data Distribution Statistic.}} The approximate distribution of the data can be estimated using statistical methods. Therefore, we can obtain an approximate distribution by computing the statistics of the local dataset, \textit{i.e.}, $P_k \sim \mathcal{N}(\mu^k, (\sigma^k)^2)$, where $\mu^k$ and $\sigma^k$ are mean and standard deviation, respectively. Specifically, to estimate such underlying distribution information of each client,  we compute the channel-wise statistics within each local dataset in client-side before the start of training:
\begin{equation}
    \mu^k = \sum_{i=1}^{n_k}\mu^k_{i} \in \mathbb{R}^{C}, \quad
 \sigma^{k} = \sum_{i=1}^{n_k}\sigma^k_{i}\in \mathbb{R}^{C}, 
\end{equation}
where $\mu^k_{i}$ and $\sigma^k_{i}$ are sample-level channel-wise statistics, and they can be computed as:
\begin{equation}
\begin{aligned}
    \mu^k_{i} &= \frac{1}{HW}\sum_{h=1}^{H}\sum_{w=1}^{W} x_{i}^{k,(h,w)}, \\ 
    \sigma^k_{i} &= \sqrt{\frac{1}{HW}\sum_{h=1}^{H}\sum_{w=1}^{W} (x_{i}^{k,(h,w)}-\mu^k_{i})^2},
\end{aligned}
\end{equation}
where $x_{i}^{k,(h,w)}$ represents the image pixel at spatial location $(h, w)$. Following this, all data distribution statistics will be sent to the server and aggregated by the server. The aggregated statistics $\{(\mu^k, \sigma^{k})\}_{k=1}^{K}$ are shared among clients.

\noindent{\textbf{Data Augmentation at Training Phase.}} After obtaining the statistical information of each client, we utilize them to augment data during training. Considering an image $x^k_{i}$, different from the normal data normalization that transforms the image according to a fixed statistic, we transform the image by randomly selecting the mean and  standard deviation from statistics $\{(\mu^k, \sigma^{k})\}_{k=1}^{K}$, which can be described as:
\begin{equation}
    \hat{x}^k_{i} = \frac{x_i^{k} - \mu^{j}}{\sigma^j}, \quad (\mu^j, \sigma^j) \sim \{(\mu^k, \sigma^{k}) \}_{k=1}^{K}. \label{eq:eq7}
\end{equation}
Notably, the statistic $(\mu^j, \sigma^j)$ will be randomly reselected for each image at each training epoch. Consequently, after multiple rounds, the number of selections surpasses the quantity of statistics by a substantial margin. This implies that each client leverages the local distribution information from all clients to augment each image, which potentially injects all local information from every client into the client-side training.  In this way, we seamlessly inject global information into augmented samples. The local model can learn the distribution information of all clients, thereby making the learned features more generalized.

\noindent{\textbf{Data Processing at Testing Phase.}}
During training, the random selection strategy integrates distribution information across all clients (\ie, local statistics). This allows the model to remain robust to variations in client-specific distributions. However, at the testing stage, communication between clients should be disabled, and each client should operate independently, without access to distribution information from others. Thus, at test time, we only apply the local statistics pertinent to each individual client for data processing. This ensures the consistency between the distribution of training and testing data. For client $k$, the above process can be formalized as follows:  
\begin{equation}
    \hat{x}^k_{i} = \frac{x_i^{k} - \mu^{k}}{\sigma^k}.
\end{equation}
The overview of two processes, \textit{i.e.}, statistic computing and data augmentation, are presented in Algorithm~\ref{alg:fedrdn}.

% \begin{table*}[!t]
% \centering
% \caption{\textbf{The number of samples in } \texttt{train}, \texttt{val} and \texttt{test} for each client from three datasets.}
% \setlength\tabcolsep{3pt}
% \renewcommand\arraystretch{1.4}{
% \begin{tabular}{lcccccccccccccccc}
% \toprule
% \multirow{2}{*}{\textbf{Samples}} & \multicolumn{4}{c}{\textbf{Office-Caltech-10}~\cite{gong2012geodesic}}
% &\multicolumn{6}{c}{\textbf{DomainNet}~\cite{peng2019moment}} & \multicolumn{6}{c}{\textbf{ProstateMRI}~\cite{liu2020ms}}
% \\

% \cmidrule(lr){2-5}
% \cmidrule(lr){6-11}
% \cmidrule(lr){12-17}
% & Amazon & Catech & DSLR & WebCam  & Clipart & Infograph & Painting & Quickdraw & Real & Sketch & BIDMC& HK & I2CVB & BMC &RUNMC & UCL\\
% \hline
% \hline

% \texttt{train} & 459 & 538 & 75 & 141&  672 & 840 &791 & 1280 & 1556  & 708 & 156 & 94 & 280 &230 & 246 &105 \\ 

% \texttt{val} &307& 360& 50& 95& 420 &525& 494& 800& 972& 442& 52 &31 &93 &76 &82 &35 \\

% \texttt{test} & 192 & 225 & 32 & 59 & 526 &657 & 619 & 1000 & 1217 & 554 & 52 &31 & 93&76& 82& 35 \\

% \bottomrule
% \end{tabular}}
% \label{tab:dataset}
% \end{table*}

\subsection{Privacy Security}

The previous input-level data augmentation method, \textit{i.e.}, FedMix~\cite{yoon2021fedmix}, shares the average images per batch, leading to the increased risk of privacy. Different from it, our method only shares the privacy irrelevant information, \textit{i.e.}, dataset-level mean and standard deviation. In addition, we can not reverse the individual image from the shared information because it is
statistical information of the whole dataset. Therefore, our method has a high level of privacy security.

\section{Experiments}

\begin{table*}[!t]
\centering
\caption{ \textbf{The test accuracy (\%) of all approaches on office-Caltech-10~\cite{gong2012geodesic} and DomainNet~\cite{peng2019moment}}. For a detailed comparison, we present the test accuracy of each client \textit{i.e.}, \textbf{Office-Caltech-10}: A(Amazon), C(Caltech), D(DSLR), W(Webcam), \textbf{DomainNet}: C(Clipart), I(Infograph), P(Painting), Q(Quickdraw), R(Real), S(Sketch), and the average result. ${\color{green}\uparrow}$ and ${\color{red}\downarrow}$ show the rise and fall of the average result before and after augmentation. We mark best results in bold. (norm.: conventional data normalization) }
\footnotesize
\setlength\tabcolsep{6pt}
\renewcommand\arraystretch{1.0}{
\begin{tabular}{lcccclccccccl}
\toprule
\multirow{2}{*}{\textbf{Method}} & \multicolumn{5}{c}{\textbf{Office-Caltech-10}}
&\multicolumn{7}{c}{\textbf{DomainNet}}
\\

\cmidrule(lr){2-6}
\cmidrule(lr){7-13}
& A & C & D & W & Avg. & C & I & P & Q & R & S & Avg. \\
\hline
\hline
FedAvg~\cite{mcmahan2017communication}    &53.12 & 44.88 & 65.62 & 86.44 & 62.51 & \textbf{50.38} & 22.83 & 36.99 &58.10 & 46.09 & 39.53 & 42.32 \\
\quad + \textit{norm}   &50.52 & 43.55 & 68.75 & 83.05 & 61.46$_{\color{coldgrey}(1.05)} 
 {\color{red}\downarrow}$ & 48.28 & 23.28 & 37.80 &54.20 & 48.97 & \textbf{41.69} & 42.37$_{\color{coldgrey}(0.05)} 
 {\color{green}\uparrow}$ \\
 \quad + \textit{FedMix}~\cite{yoon2021fedmix}   &49.47 & 41.77 & 75.00 & 88.13 & 63.59$_{\color{coldgrey}(1.08)} {\color{green}\uparrow}$ & 48.66 & \textbf{23.43} & 38.12 &55.10 & 49.46 & 41.33 & 42.68$_{\color{coldgrey}(0.36)} {\color{green}\uparrow}$ \\
\rowcolor[HTML]{EFEFEF}  \quad  + \textbf{\textit{FedRDN}}   &\textbf{60.93} & \textbf{45.77} & \textbf{84.37} & \textbf{88.13} & \textbf{69.80}$_{\color{coldgrey}(7.29)} {\color{green}\uparrow}$ & 48.85 & 22.67 & \textbf{39.41} &\textbf{60.30} & \textbf{49.46} & 40.61 & \textbf{43.55}$_{\color{coldgrey}(1.23)} {\color{green}\uparrow}$ \\
\hline
FedProx~\cite{li2020federated}&53.12 & 45.33 & 62.50 & 86.44 & 61.84 & \textbf{52.66} & 23.89 & 35.21 &56.70 & 46.75 & 41.87 & 42.85 \\
\quad+ \textit{norm}   &51.04 & \textbf{45.77} & 68.75 & 84.74 & 62.57$_{\color{coldgrey}(0.73)} {\color{green}\uparrow}$ & 47.14 & 24.35 & 34.57 &59.60 & 44.86 & 38.98 & 41.58$_{\color{coldgrey}(1.27)} {\color{red}\downarrow}$ \\
\quad+ \textit{FedMix}~\cite{yoon2021fedmix}   &47.39 & 38.66 & 78.12 & \textbf{91.52} & 63.92$_{\color{coldgrey}(2.08)} {\color{green}\uparrow}$ & 47.90 & 22.37 & 37.31 &53.90 & 48.47 & \textbf{43.14} & 42.18$_{\color{coldgrey}(0.67)} {\color{red}\downarrow}$\\
\rowcolor[HTML]{EFEFEF} \quad+ \textbf{\textit{FedRDN}}     &\textbf{61.45} & 44.88 & \textbf{84.37} & 88.13 & \textbf{69.71}$_{\color{coldgrey}(7.87)} {\color{green}\uparrow}$ & 50.57 & \textbf{24.96} & \textbf{38.77} &\textbf{61.20} & \textbf{51.35} & 40.97 & \textbf{44.63}$_{\color{coldgrey}(1.78)} {\color{green}\uparrow}$ \\
\hline
FedNova~\cite{wang2020tackling}     & 50.00 & 42.22 & 62.50 & 88.13 & 60.71 & \textbf{51.71} & 23.74 & 38.77 & 56.20 & 45.52 &38.44 & 42.39 \\
\quad+ \textit{norm} & 52.08 &
\textbf{45.33} & 68.75 & 86.44 & 63.15$_{\color{coldgrey}(2.44)} {\color{green}\uparrow}$ & 49.23 & \textbf{24.35} & 34.24 & 55.80 & 45.52 &42.23 & 41.90$_{\color{coldgrey}(0.49)} {\color{red}\downarrow}$   \\
\quad+ \textit{FedMix}~\cite{yoon2021fedmix}  &48.95 &42.66 &78.12 &83.05 &63.20$_{\color{coldgrey}(2.49)} {\color{green}\uparrow}$ & 47.90 &24.04 &36.67 &\textbf{59.10} &46.67 &\textbf{42.41}  &42.80$_{\color{coldgrey}(0.41)} {\color{green}\uparrow}$ \\ 
\rowcolor[HTML]{EFEFEF} \quad+ \textbf{\textit{FedRDN}}   & \textbf{63.02} &
41.33 & \textbf{84.37} & \textbf{89.83} & \textbf{69.63}$_{\color{coldgrey}(8.71)} {\color{green}\uparrow}$ & 50.57 & 23.43 & \textbf{40.22} & 57.30 & \textbf{51.84} &40.43 & \textbf{43.96}$_{\color{coldgrey}(1.57)} {\color{green}\uparrow}$ \\
\hline
Scaffold~\cite{karimireddy2020scaffold}   & 52.60 & \textbf{42.66} & 53.12 & 81.35 & 57.43 & 46.95 &22.83 &34.57 &46.50 &47.00 &40.97 &39.80 \\
 \quad+ \textit{norm} & 46.87 & 40.00 & 59.37 &86.44 &58.17$_{\color{coldgrey}(0.74)} {\color{green}\uparrow}$&47.33 &22.83 &33.11 &58.30 &46.01 &\textbf{42.05} &41.61$_{\color{coldgrey}(1.81)} {\color{green}\uparrow}$ \\
  \quad+ \textit{FedMix}~\cite{yoon2021fedmix} & 52.08 &40.88 &75.00 &\textbf{89.83} &64.45$_{\color{coldgrey}(7.02)} {\color{green}\uparrow}$& 45.24 &23.28 &34.73 &47.50 &44.78 &40.97 &39.42$_{\color{coldgrey}(0.38)} {\color{red}\downarrow}$ \\
\rowcolor[HTML]{EFEFEF} \quad+ \textbf{\textit{FedRDN}}   & \textbf{65.10} & 41.77 & \textbf{81.25} &86.44 &\textbf{68.64}$_{\color{coldgrey}(11.21)} {\color{green}\uparrow}$ & \textbf{51.52} &\textbf{23.89} &\textbf{37.96} &\textbf{56.20} &\textbf{48.97} &38.80 &\textbf{42.89}$_{\color{coldgrey}(3.09)} {\color{green}\uparrow}$ \\
\hline
FedAvgM~\cite{hsu2019measuring} & 48.43 & \textbf{45.33} & 62.50 & 83.05 & 59.83& 45.81 & 22.52 & 37.96 & 50.10 & 48.23 & \textbf{41.87}& 41.08\\
\quad+ \textit{norm}  & 51.04 & 44.88 & 62.50 & 86.44 & 61.21$_{\color{coldgrey}(1.38)} {\color{green}\uparrow}$& 46.95 & 24.96 & 35.86 & 49.70 & 45.43 & 40.43& 40.55$_{\color{coldgrey}(0.53)} {\color{red}\downarrow}$ \\
\quad+ \textit{FedMix}~\cite{yoon2021fedmix}  & 50.00 &41.77 &65.62 &83.05 &60.11$_{\color{coldgrey}(0.29)} {\color{green}\uparrow}$ & \textbf{48.28} & 
\textbf{25.87} & 40.06 & 51.50 & 48.56 & 38.62& 42.15$_{\color{coldgrey}(1.07)} {\color{green}\uparrow}$ \\

\rowcolor[HTML]{EFEFEF}  \quad+ \textbf{\textit{FedRDN}}    & \textbf{62.50} & 43.11 & \textbf{84.37} & \textbf{88.13} & \textbf{69.53}$_{\color{coldgrey}(9.70)} {\color{green}\uparrow}$& 48.09 & 22.98 & \textbf{41.03} & \textbf{63.80} & \textbf{49.79} & 38.08& \textbf{43.96}$_{\color{coldgrey}(2.88)} {\color{green}\uparrow}$\\

\hline
FedBN~\cite{li2021fedbn} & 67.18 & \textbf{44.00} & \textbf{84.97} & 86.44 & 70.65& 49.23 & 24.96 & 36.32 & 63.60 & 47.74 & 39.53& 43.56\\
 \quad+ \textit{norm}  & 64.84 & 41.07 & 84.11 & 87.87 & 69.47$_{\color{coldgrey}(1.18)} {\color{red}\downarrow}$& 48.47 & 23.81 & 36.67 & \textbf{63.76} & \textbf{49.71} & 40.79& 43.88$_{\color{coldgrey}(0.32)} {\color{green}\uparrow}$ \\
\quad+ \textit{FedMix}~\cite{yoon2021fedmix}  & 65.62 &44.88 &84.37 &88.45 &70.83$_{\color{coldgrey}(0.18)} {\color{green}\uparrow}$ & 49.61 & \textbf{25.72} & \textbf{38.44} & 57.30 & 46.01 & 38.08& 42.53$_{\color{coldgrey}(1.03)} {\color{red}\downarrow}$ \\

\rowcolor[HTML]{EFEFEF} \quad+ \textbf{\textit{FedRDN}}    & \textbf{67.74} & 43.11 & 84.37 & \textbf{89.35} & \textbf{71.14}$_{\color{coldgrey}(0.49)} {\color{green}\uparrow}$& \textbf{50.76} & 25.26 & 37.15 & 61.98 & 48.97 & \textbf{43.14}& \textbf{44.54}$_{\color{coldgrey}(0.98)} {\color{green}\uparrow}$\\

\hline
FedProto~\cite{tan2022fedproto}  & 55.72 & 44.44 & 68.75 & 86.44 & 63.84 & 48.28 & \textbf{25.11} & 35.86 & 51.30 & 43.79 & 37.18 & 40.25 \\
 \quad+ \textit{norm}  & 53.64 & 44.88 & 56.25 & 86.44 & 60.30$_{\color{coldgrey}(3.54)} {\color{red}\downarrow}$ & 45.81 & 23.43 & 35.70 & \textbf{58.30} &45.27 &40.79 &41.55$_{\color{coldgrey}(1.30)} {\color{green}\uparrow}$ \\
 \quad+ \textit{FedMix}~\cite{yoon2021fedmix}  & 53.64 &41.77 &84.37 &88.13 &66.98$_{\color{coldgrey}(3.14)} {\color{green}\uparrow}$ & 47.33 & 23.43 & 37.47 & 52.70 &44.94 &\textbf{42.41} &41.38$_{\color{coldgrey}(1.13)} {\color{green}\uparrow}$\\
\rowcolor[HTML]{EFEFEF}\quad+ \textbf{\textit{FedRDN}}    & \textbf{66.14} & \textbf{46.22} & \textbf{84.37} & \textbf{89.83} & \textbf{71.64}$_{\color{coldgrey}(7.80)} {\color{green}\uparrow}$& \textbf{49.42} & 22.37 & \textbf{41.51} & 57.90 & \textbf{51.43} & 38.44 & \textbf{43.51}$_{\color{coldgrey}(3.26)} {\color{green}\uparrow}$\\
\hline
FedFA~\cite{zhou2023fedfa}   & 60.93 &48.44 &81.25 &89.83 &70.11 & 48.09 &23.74 &\textbf{39.58} &\textbf{64.20} &48.06 &\textbf{43.14} &44.47 \\
  \quad+ \textit{norm} & 60.93 &\textbf{50.66} &81.25 &84.74 &69.40$_{\color{coldgrey}(0.71)} {\color{red}\downarrow}$ & 49.04 &24.20 &38.28 &60.70 &45.93 &42.59 &43.46$_{\color{coldgrey}(1.01)} {\color{red}\downarrow}$ \\
   \quad+ \textit{FedMix}~\cite{yoon2021fedmix}  & 56.25 &48.44 &84.37 &88.13 &69.30$_{\color{coldgrey}(0.81)} {\color{red}\downarrow}$ & 45.81 & 22.52& 33.11& 51.10&43.13& 37.90 &38.93$_{\color{coldgrey}(5.54)} {\color{red}\downarrow}$ \\
\rowcolor[HTML]{EFEFEF}   \quad+ \textbf{\textit{FedRDN}}   & \textbf{62.50} &48.88 &\textbf{90.62} &\textbf{91.52} &\textbf{73.38}$_{\color{coldgrey}(3.27)} {\color{green}\uparrow}$& \textbf{52.85} &\textbf{24.50} &37.80 &61.90 &\textbf{50.45} &42.59 &\textbf{45.01}$_{\color{coldgrey}(0.54)} {\color{green}\uparrow}$ \\

\bottomrule
\end{tabular}}
\label{tab:merge}
\end{table*}

\subsection{Experimental Setup}
\noindent{\textbf{Datasets.}} We conduct extensive experiments on three real-world datasets: \textbf{Office-Caltech-10}~\cite{gong2012geodesic}, \textbf{DomainNet}~\cite{peng2019moment}, and \textbf{ProstateMRI}~\cite{liu2020ms}, which are widely used in the feature distribution skewed FL scenario~\cite{li2021fedbn, zhou2023fedfa, jiang2022harmofl}. There are two different tasks including image classification (Office-Caltech-10 and DomainNet) and medical image segmentation (ProstateMRI). Following previous work~\cite{li2021fedbn,zhou2023fedfa}, we employ the subsets as clients when conducting experiments on each dataset.  

\noindent{\textbf{Baselines.}} To demonstrate the effectiveness of our method, we build four different data augmentation flows: \ding{182} \textbf{\textit{base}}: one flow has some basic data augmentation techniques like random flipping, \ding{183} \textbf{\textit{+norm}:} one flow adds the conventional normalization technique into \textit{base}, \ding{184} \textbf{\textit{+FedMix}:} another flow adds the FedMix~\cite{yoon2021fedmix} data augmentation technique into \textit{base}, and \ding{185} \textbf{\textit{+FedRDN}:} the rest one integrates our proposed augmentation method with  \textit{base}. Since it is infeasible to deploy FedMix into segmentation tasks, we only utilize it for image classification tasks. Following, we integrate them into different typical FL methods. In detail, we employ eight state-of-the-art FL methods to demonstrate the generalizability of our method, including \textbf{FedAvg}~\cite{mcmahan2017communication}, \textbf{FedAvgM}~\cite{hsu2019measuring}, \textbf{FedProx}~\cite{li2020federated}, 
\textbf{Scaffold}~\cite{karimireddy2020scaffold},
\textbf{FedNova}~\cite{wang2020tackling}, \textbf{FedBN}~\cite{li2021fedbn},
\textbf{FedProto}~\cite{tan2022fedproto}, and \textbf{FedFA}~\cite{zhou2023fedfa} for validation of image classification task. Moreover, we select five of them which are general for different tasks to validate the effectiveness of our method on medical image segmentation tasks. To quantitatively evaluate the performance, we utilize the top-1 accuracy for image classification while the medical segmentation is evaluated with the Dice coefficient. 

% \noindent{\textbf{Network Architecture.}} Following previous work~\cite{li2021fedbn, zhou2023fedfa}, we employ the Alex-Net~\cite{krizhevsky2017imagenet} as the image classification model and the U-Net~\cite{ronneberger2015u} as the medical image segmentation model.

\begin{figure*}[!t]
\centering
  \includegraphics[width=0.96\textwidth]{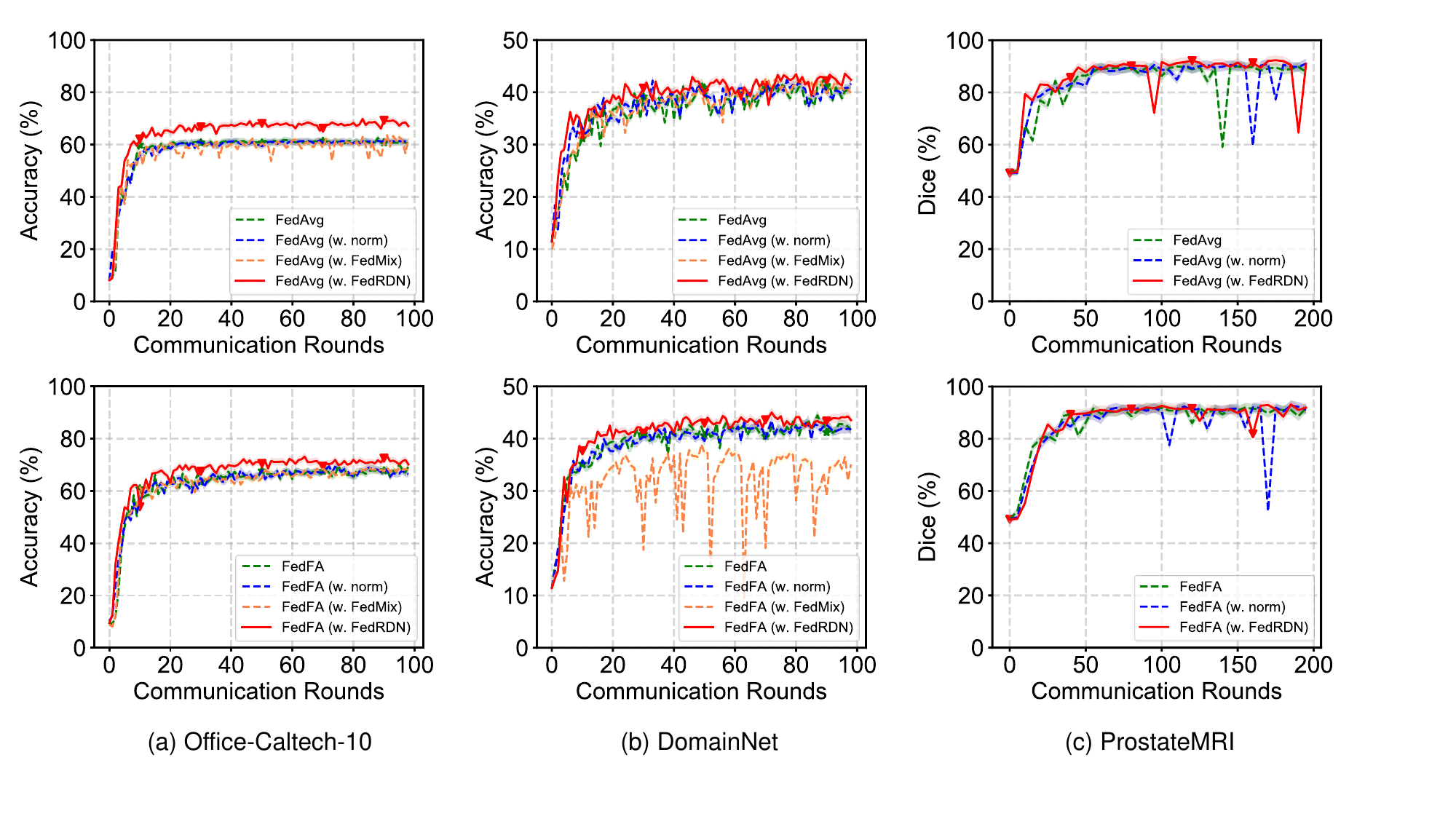}
  \caption{ \textbf{Illustration of test performance versus communication rounds} on (a) Office-Caltech-10~\cite{gong2012geodesic}, (b) DomainNet~\cite{peng2019moment}, and (c) ProstateMRI~\cite{liu2020ms}. }
  \label{fig:comm}
\end{figure*}

\noindent{\textbf{Implementation Details.}} All methods are implemented by PyTorch, and we conduct all experiments on a single NVIDIA GTX 1080Ti GPU with 11GB of memory.   Following previous work~\cite{li2021fedbn, zhou2023fedfa}, we employ the Alex-Net~\cite{krizhevsky2017imagenet} as the image classification model and the U-Net~\cite{ronneberger2015u} as the medical image segmentation model. We use cross-entropy loss for image classification and Dice loss for medical image segmentation. The batch size is 32  for two image classification datasets and 16 for the ProstateMRI dataset. We adopt the SGD optimizer with a learning rate of 0.01 and weight decay of 1e-5 for image classification datasets, and the Adam optimizer with a learning rate of 1e-3 and weight decay of 1e-4 for ProstateMRI. Furthermore, we run 100 communication rounds on image classification tasks while the number of rounds for the medical segmentation task is 200, and each round has 5 epochs of local training. More importantly, for a fair comparison, we train all methods in the same environment and ensure that all methods have converged. Due to page limitation, we present additional experimental details and results in \textbf{\textit{Supplementary Material}}.

\begin{table}[!t]
\centering
\caption{ \textbf{The dice score (\%) of all approaches on ProstateMRI~\cite{liu2020ms}}. For a detailed comparison, we present the test results of six clients: BIDMC, HK, I2CVB, BMC, RUNMC, UCL, and the average result. ${\color{green}\uparrow}$ and ${\color{red}\downarrow}$ show the rise and fall of the average result before and after augmentation. We mark best results in bold. (norm.: conventional data normalization)}
\footnotesize
\setlength\tabcolsep{1.5pt}
\renewcommand\arraystretch{1.0}{
\begin{tabular}{lccccccl}
\toprule
\multirow{2}{*}{\textbf{Method}}
&\multicolumn{7}{c}{\textbf{ProstateMRI}}
\\
\cmidrule(lr){2-8} 
&BIDMC &HK & I2CVB & BMC & RUNMC & UCL & Avg. \\

\hline
\hline
FedAvg~\cite{mcmahan2017communication}     & 84.16 & \textbf{94.51} & 94.60 &88.43 & 92.78 & 85.64 & 90.02\\
\quad+ \textit{norm}   & 86.20 & 92.53 & \textbf{94.74} &89.85 & 92.18 & 87.91 & 90.57$_{\color{coldgrey}(0.55)} {\color{green}\uparrow}$\\
 \rowcolor[HTML]{EFEFEF} \quad+ \textbf{\textit{FedRDN}}     & \textbf{89.34} & 94.41 & 93.85 &\textbf{91.46} & \textbf{94.19} & \textbf{90.65} & \textbf{92.32}$_{\color{coldgrey}(2.30)} {\color{green}\uparrow}$\\
\hline
FedProx~\cite{li2020federated}   & 84.47 & \textbf{94.60} &94.87 & 90.29 & 92.72 & 86.60 & 90.59 \\
 \quad+ \textit{norm} & 84.47 & 94.48 &95.06 & 88.79 & 92.90 & 85.35 & 90.18$_{\color{coldgrey}(0.41)} {\color{red}\downarrow}$ \\
\rowcolor[HTML]{EFEFEF} \quad+ \textbf{\textit{FedRDN}} & \textbf{88.87} & 94.19 &\textbf{95.09} & \textbf{90.99} & \textbf{93.03} & \textbf{89.17} & \textbf{91.89}$_{\color{coldgrey}(1.30)} {\color{green}\uparrow}$ \\
\hline
FedAvgM~\cite{hsu2019measuring} & 87.02 &94.32 & 94.29 & \textbf{91.35} & 92.83 & 86.75 &91.09 \\
\quad+ \textit{norm} &
\textbf{89.05} & 93.59 &94.75 &89.93 &\textbf{93.52}&88.22 &91.51$_{\color{coldgrey}(0.42)} {\color{green}\uparrow}$ \\

\rowcolor[HTML]{EFEFEF} \quad+ \textbf{\textit{FedRDN}} &
{88.37} & \textbf{94.67} &\textbf{95.40} &90.40 &93.28&\textbf{88.39} &\textbf{91.75}$_{\color{coldgrey}(0.66)} {\color{green}\uparrow}$ \\

\hline
FedBN~\cite{li2021fedbn}   &86.42 &\textbf{94.45}&95.27&90.96&93.13&87.48&91.28 \\
 \quad+ \textit{norm} &87.45 &93.01 &95.44 &90.16 &93.22 &86.88 &91.02$_{\color{coldgrey}(0.26)} {\color{red}\downarrow}$ \\
\rowcolor[HTML]{EFEFEF} \quad+ \textbf{\textit{FedRDN}} &\textbf{89.47} &93.61 &\textbf{95.65} &\textbf{90.99} &\textbf{93.26} &\textbf{87.72} &\textbf{91.78}$_{\color{coldgrey}(0.50)} {\color{green}\uparrow}$ \\

\hline
FedFA~\cite{zhou2023fedfa}   &89.18 &92.77&94.18&\textbf{92.62}&93.63&89.04&91.90 \\
\quad+ \textit{norm} &89.12 &94.40 &95.22 &91.95 &93.42 &89.28 &92.23$_{\color{coldgrey}(0.33)} {\color{green}\uparrow}$ \\
\rowcolor[HTML]{EFEFEF} \quad+ \textbf{\textit{FedRDN}} &\textbf{91.81} &\textbf{94.65} &\textbf{95.67} &92.37 &\textbf{94.33} &\textbf{90.19} &\textbf{93.14}$_{\color{coldgrey}(1.24)} {\color{green}\uparrow}$ \\

\bottomrule
\end{tabular}}
\label{tab:table3}
\end{table}

\subsection{Main Results}

In this section, we present the overall results on Office-Caltech-10 and DomainNet in Table~\ref{tab:merge} and ProstateMRI in Table~\ref{tab:table3}. For a detailed comparison, we present the test accuracy of each client and the average result.

~\textbf{\textit{All FL methods yield significant improvements combined with FedRDN consistently over three datasets.}} As we can see, FedRDN leads to consistent performance improvement for all FL baselines across three benchmarks compared with using the basic data augmentation flow. The improvements of FedRDN can be large as \textbf{11.21\%} on Office-Caltech-10, \textbf{3.26\%} on DomainNet, and \textbf{1.37\%} on ProstateMRI, respectively. Especially for FedFA, the state-of-the-art FL method for feature distribution skewed FL setting, can still gain improvements, \textit{e.g.}, \textbf{3.27\%} on Office-Caltech-10. This indicates that input-level augmentation and feature-level augmentation are not contradictory and can be used simultaneously. Due to the utilization of personalized batch normalization layers, the advancements achieved with FedBN might not be as pronounced as with other methods. However, it's crucial to note that there are still modest improvements present. Moreover, some weaker FL methods can even achieve better performance than others when using FedRDN. For example, after using FedRDN, the accuracy of FedAvg can be significantly higher than all other FL methods except FedFA, and FedProto can be even higher than FedFA. The above results demonstrate the effectiveness of the data-level solution, which can effectively mitigate feature shift. Besides, compared with other heterogeneous FL methods, our method has a stronger flexibility and generalization ability. 

\textbf{\textit{FedRDN is superior to other input-level data augmentation techniques.}} As shown in Table~\ref{tab:merge} and ~\ref{tab:table3}, FedRDN shows leading performance compared with conventional data normalization and FedMix, a previous input-level data augmentation technique. Besides, these two augmentation techniques can even decrease the performance of the method in several cases, while FedRDN can achieve consistent improvements. For instance, FedAvg and FedProto yield a drop as large as \textbf{1.05\%} and  \textbf{3.54\%} with conventional data normalization on Office-Caltech-10, respectively. FedProx and FedFA show a drop as \textbf{0.67\%} and \textbf{5.54\%} on DomainNet, respectively, when they are combined with FedMix. The above results demonstrated the effectiveness of FedRDN, and it has stronger compatibility compared with other data augmentation methods.

\begin{table}[!t]
\centering
\caption{ \textbf{Generalization performance of local model} on Office-Caltech-10~\cite{gong2012geodesic}. We mark the best result in bold.}
\setlength\tabcolsep{2pt}
\footnotesize
\renewcommand\arraystretch{1.0}{{\begin{tabular}{l|l|c|c|c|c}
\toprule
Source-site& Target-site & FedAvg~\cite{mcmahan2017communication} &  + \textit{norm} &  + \textit{FedMix}~\cite{yoon2021fedmix} & + \textit{\textbf{FedRDN}} \\
\hline
\hline

\multirow{3}{*}{Amazon} & Caltech & 48.00 & 48.09 & 42.22 & \textbf{48.88} \\
& DSLR & 46.87 & 44.41 & 59.37 & \textbf{81.25} \\
& Webcam & 77.96 & 77.96 & \textbf{84.74} & 83.05 \\
\hline
\multirow{3}{*}{Caltech} & Amazon & 58.33 & 58.85 & 51.56 & \textbf{63.02} \\
& DSLR & 68.75 & 68.75 & 75.00 & \textbf{84.37} \\
& Webcam & 83.05 & 83.05 & \textbf{91.52} & 86.35 \\
\hline
\multirow{3}{*}{DSLR} & Amazon & 42.70 & 41.14 & 33.85 & \textbf{60.93} \\
& Caltech & 33.33 & 35.11 & 35.55 & \textbf{35.56} \\
& Webcam & 79.66 & 77.96 & 74.57 & \textbf{84.74} \\
\hline
\multirow{3}{*}{Webcam} & Amazon & 53.64 & 55.20 & 47.91 & \textbf{65.10} \\
& Caltech & 41.33 & 43.55 & 40.00 & \textbf{48.00} \\
& DSLR & 68.75 & 68.75 & 78.12 & \textbf{87.50} \\

\bottomrule
\end{tabular}}}
\label{tab:table5}
\end{table}

\subsection{Communication Efficiency}

\noindent{\textbf{Convergence.}} To explore the impact of various data augmentation techniques on the convergence,  we draw the test performance curve of FedAvg and a state-of-the-art FL method, \textit{i.e.}, FedFA, with different communication rounds on three datasets as shown in Fig.~\ref{fig:comm}. Apparently, FedRDN will not introduce any negative impact on the convergence of the method and even yield a faster convergence at the early training stage (0 $\sim$ 10 rounds) in some cases. As the training goes on, FedRDN achieves a more optimal solution. Besides, compared with other methods, the convergence curves of FedRDN are more stable.

\noindent{\textbf{Communication Cost.}} In addition to the existing communication overhead of FL methods, the additional communication cost in FedRDN is only for statistical information. The dimension of the statistic is so small (mean and standard deviation are $\mathbb{R}^3$ for RGB images), that the increased communication cost can be neglected. This is much different from the FedMix, which needs to share the average images per batch. The increased communication cost of FedMix is as large as 156MB on Office-Caltech-10 and 567MB on DomainNet while the batch size of averaged images is 5, even larger than the size of model parameters.

\subsection{Cross-site Generalization performance}
As stated before, FedRDN augments the data with luxuriant distribution from all clients to learn the more generalized feature representation. Therefore, we further explore the generalization performance of local models by cross-site evaluation, and the results are presented in Table~\ref{tab:table5}. As we can see, All local models of FedRDN yield a consistent improvement over FedAvg, and our method shows better generalization performance compared to other methods. The above results demonstrate that FedRDN can effectively improve the generalization of local feature representation, thereby mitigating feature shift across different clients. This is the reason why our method works.

\begin{table}[!t]
\centering
\caption{ \textbf{The performance of FedRDN over FedRDN-V} on three datasets.}
\setlength\tabcolsep{3pt}
\footnotesize
\renewcommand\arraystretch{1.0}{
\begin{tabular}{lccc}
\toprule
\textbf{Method}&\textbf{Office-Caltech-10}& \textbf{DomainNet}&\textbf{ProstateMRI}\\
\hline
\hline
FedAvg~\cite{mcmahan2017communication}  & $62.51$ & $42.32$ & $90.02$ \\
\quad + \textit{FedRDN-V} & $61.46$ & $42.99$ & $91.14$ \\

\rowcolor[HTML]{EFEFEF} \quad + \textbf{\textit{FedRDN} (Ours)} & $\textbf{69.80}$ & $\textbf{43.55}$ & $\textbf{92.32}$ \\

\bottomrule
\end{tabular}}
\label{tab:table4}
\end{table}

\begin{figure}[t]
  \centering
  \includegraphics[width=0.48\textwidth]{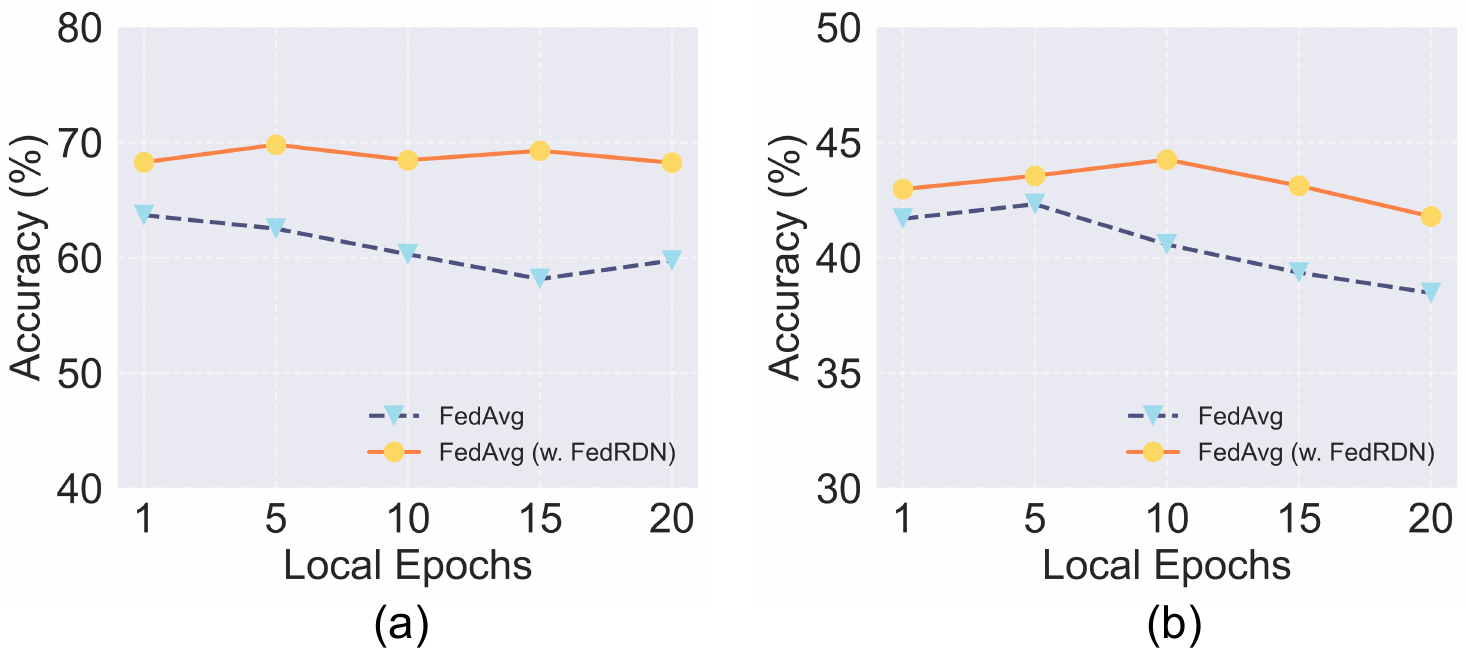}
  \caption{\textbf{Illustration of test performance versus local epochs} on (a) Office-Caltech-10~\cite{gong2012geodesic} and (b) DomainNet~\cite{peng2019moment}.}
  \label{fig:local}
\end{figure}

\subsection{Analytical Studies}

\noindent{\textbf{FedRDN vs. FedRDN-V.}}  To deeply explore FedRDN, we develop a variant, FedRDN-V. Instead of randomly transforming, it transforms the images with the average mean $\hat{u}$ and standard deviation $\hat{\sigma}$ of all clients during training and testing phases:
\begin{equation}
    \hat{u} = \sum_{k=1}^K \mu_k, \quad \hat{\sigma} = \sum_{k=1}^K \sigma_k.
\end{equation}
 This degenerates into traditional data normalization with the hyper-parameter $\hat{u}$ and  $\hat{\sigma}$. The results of the comparison over three datasets are presented in Table~\ref{tab:table4}. Apparently, FedRDN-V yields a significant drop compared with our method. This indicates the effectiveness of our method is not from the traditional data normalization but augmenting samples with the information from the multiple real local distributions. By this, each local model will be more generalized instead of biasing in skewed underlying distributions.

\begin{figure}[!t]
\centering
  \includegraphics[width=0.48\textwidth]{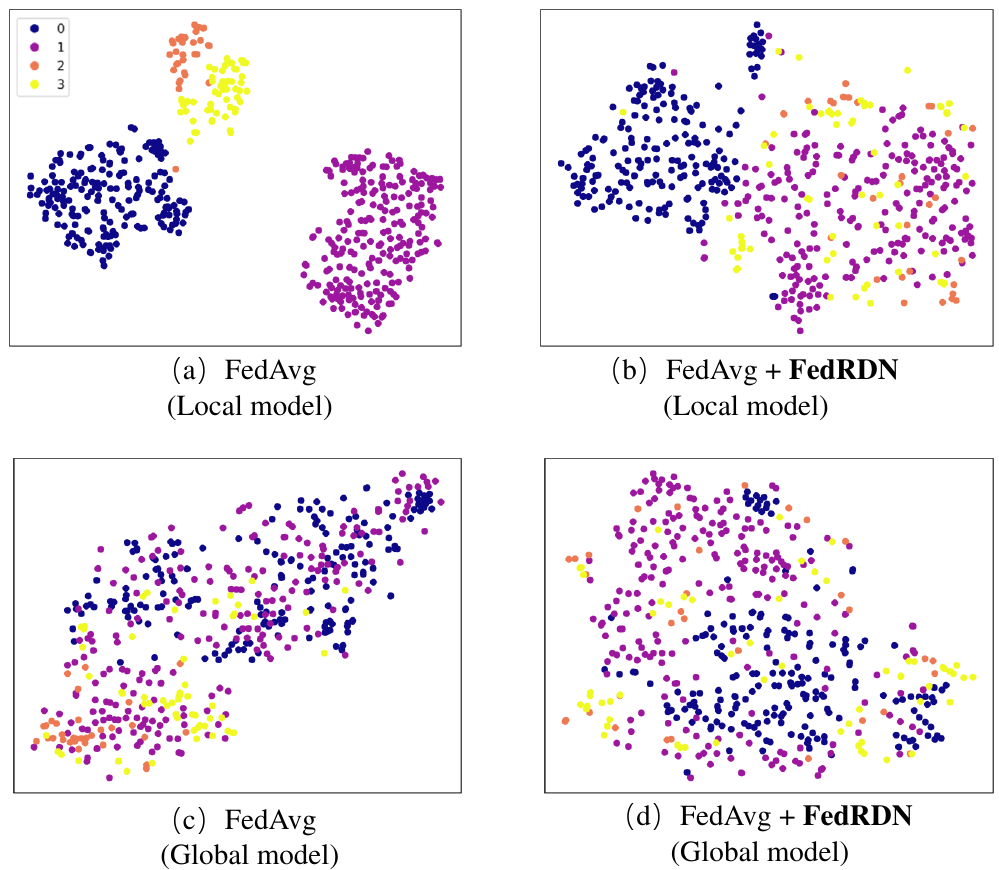}
  \caption{\textbf{T-SNE visualization of features } on Office-Caltech-10~\cite{gong2012geodesic}. The T-SNE is conducted on the test sample features of four clients. We use different colors to mark features from different clients.}
  \label{fig:tsne}
\end{figure}
\noindent{\textbf{Robust to Local Epochs.}} To explore the robustness of FedRDN for different local epochs, we tune the local epochs from the \{1, 5, 10, 15, 20\} and evaluate the performance of the trained model. The results are presented in Fig.~\ref{fig:local}. Generally, more epochs of local training will increase the discrepancy under non-IID data, leading to slower convergence, which degrades the performance at the same communication rounds. The result of FedAvg validates this. By contrast, our method can obtain consistent improvements with different settings of local epochs. Moreover, our approach has stable performance across different local epoch settings due to effectively addressing feature shift.

\noindent{\textbf{Varying Architectures.}} To validate the robustness of our method for different network architectures, we conduct additional experiments for different input-level augmentation methods using ResNet-18~\cite{he2016deep} on Office-Caltech-10. We only change the network and do not alter any other experimental setup. The results are presented in Table~\ref{tab:vary}.  Due to the small size of the dataset, the performance of ResNet-18 is not as good as that of AlexNet. This is consistent with the findings reported in~\cite{zhuang2023normalization}. However, the results still show the consistent improvements obtained through our proposed augmentation approach, indicating its strong robustness and adaptability for different network architectures.

\begin{table}[!t]
\centering
\caption{ \textbf{Accuracy(\%) of input-level data augmentation methods with varying architectures} on Office-Caltech-10~\cite{gong2012geodesic}.}
\setlength\tabcolsep{3pt}
\footnotesize
\renewcommand\arraystretch{1.0}{
\begin{tabular}{l|c|c|c|c}
\toprule
Network & FedAvg~\cite{mcmahan2017communication} &  + \textit{norm} &  + \textit{FedMix}~\cite{yoon2021fedmix} &  + \textit{\textbf{FedRDN}} \\
\hline
\hline
AlexNet & 62.51 & 61.46 & 63.59 & \textbf{69.80} \\
ResNet-18 & 43.74 &	43.60 &	53.81 &	\textbf{56.70} \\
\bottomrule
\end{tabular}}
\label{tab:vary}
\end{table}

\noindent{\textbf{Feature Distribution.}} To yield more insights about FedRDN, we utilize T-SNE~\cite{van2008visualizing} 
, a popular analysis tool, to visualize the distribution of the features outputted by the local model (Fig.~\ref{fig:tsne} (a) and (b)) and global model (Fig.~\ref{fig:tsne} (c) and (d)) before and after augmentation.  Specifically, we visualize the features of test samples for each client and mark them with different colors: 0 (Amazon), 1 (Caltech), 2 (DSLR), and 3 (Webcam). Due to the different data distributions among these clients, their local feature distributions exhibit a shift (Fig.~\ref{fig:tsne} (a)). This makes it challenging for the global model to learn consistent feature distributions across different clients. Intuitively, as shown in Fig.~\ref{fig:tsne} (c), the features of client 2 (DSLR) are clustered in the bottom-left corner, showing a noticeable shift from the features of other clients. In contrast, our method significantly improves the feature generalization of local models, narrowing the distance between different local feature distributions (Fig.~\ref{fig:tsne} (b)). This benefits the global model to learn a more consistent feature distribution. As shown in Fig.~\ref{fig:tsne} (d), the global features of all four clients are uniformly distributed, indicating that the features from these clients are now in a shared feature space. The above results reveal the working principle of FedRDN and further confirm its effectiveness in mitigating feature shift.

\section{Conclusion}
In this paper, we focus on addressing the feature distribution skew in FL. Different from previous insights for this problem, we try to solve this challenge from the input-level data. The proposed novel data augmentation technique, FedRDN, is a plug-and-play component that can be easily integrated into the data augmentation flow, thereby effectively mitigating feature shift. Our extensive experiments show that FedRDN can further improve the performance of various state-of-the-art FL methods across three datasets, which demonstrates the effectiveness, compatibility, and generalizability of our method. 

\section{Limitation} This work primarily focuses on visual tasks, where data statistics are privacy-agnostic information, as they only capture the distribution information instead of individual-level information. Besides, we mainly demonstrated the effectiveness of FedRDN in the feature distribution skewed FL scenario by the empirical analysis. In the future, we plan to undertake an in-depth theoretical analysis to further enhance the understanding and explanation of our method. Considering the effectiveness, compatibility, and generalizability of FedRDN, we believe this work contributes substantively to advancing the development of FL. 

{
    \small
    \bibliographystyle{ieeenat_fullname}
    \bibliography{main}
}

 \clearpage
\setcounter{page}{1}
\maketitlesupplementary
\appendix

\begin{table*}[hb]
\centering
\caption{\textbf{The number of samples in } \texttt{train}, \texttt{val} and \texttt{test} for each client on three datasets.}
\footnotesize
\setlength\tabcolsep{2pt}
\renewcommand\arraystretch{1.4}{
\begin{tabular}{lcccccccccccccccc}
\toprule
\multirow{2}{*}{\textbf{Samples}} & \multicolumn{4}{c}{\textbf{Office-Caltech-10}~\cite{gong2012geodesic}}
&\multicolumn{6}{c}{\textbf{DomainNet}~\cite{peng2019moment}} & \multicolumn{6}{c}{\textbf{ProstateMRI}~\cite{liu2020ms}}
\\

\cmidrule(lr){2-5}
\cmidrule(lr){6-11}
\cmidrule(lr){12-17}
& Amazon & Catech & DSLR & WebCam  & Clipart & Infograph & Painting & Quickdraw & Real & Sketch & BIDMC& HK & I2CVB & BMC &RUNMC & UCL\\
\hline
\hline

\texttt{train} & 459 & 538 & 75 & 141&  672 & 840 &791 & 1280 & 1556  & 708 & 156 & 94 & 280 &230 & 246 &105 \\ 

\texttt{val} &307& 360& 50& 95& 420 &525& 494& 800& 972& 442& 52 &31 &93 &76 &82 &35 \\

\texttt{test} & 192 & 225 & 32 & 59 & 526 &657 & 619 & 1000 & 1217 & 554 & 52 &31 & 93&76& 82& 35 \\

\bottomrule
\end{tabular}}
\label{tab:dataset}
\end{table*}

\section{More Details} \label{Sec:A_A}

\subsection{Details of Dataset}

There are 4 subsets for Office-Caltech-10~\cite{gong2012geodesic} and 6 subsets for DomainNet~\cite{peng2019moment} and ProstateMRI~\cite{liu2020ms}. These subsets from different domains, incurring feature distribution skew.  Following previous work~\cite{li2021fedbn,zhou2023fedfa}, we employ the subsets as clients when conducting experiments on each dataset. Therefore, the number of clients for Office-Caltech-10, DomainNet, and ProstateMRI is 4, 4 and 6, respectively. Both Office-Caltech-10 and DomainNet consist of 10 classes each. ProstateMRI, on the other hand, is a binary segmentation task involving lesions and background. The number of samples in \texttt{train}, \texttt{val} and \texttt{test} for each client can be seen in Table~\ref{tab:dataset}. 

\subsection{Details of Baselines}

There are some important hyper-parameters for some FL methods. For instance, the FedProx and FedProto have $\mu$ to control the contribution of an additional loss function. We empirically set $\mu$ to 0.001 for FedProx and 1 for FedProto on all datasets. Besides, FedAvgM has a momentum hyper-parameter to control the momentum update of the global model parameters, which is set to 0.9 for two image classification datasets and 0.01 for ProstateMRI. FedMix also has two hyper-parameters, \textit{i.e.}, the batch size of mean images and $\lambda$ to control images fusing. We adopted the reported configuration from the original paper. The batch size is set to 5 and $\lambda$ is set to 0.1.

\section{More Experiments} \label{Sec:A_B}

\subsection{Stability of Our method}

In this section, we further evaluate the stability of our method. To this end, we conducted two additional independent trials with different random seeds and reported the mean and standard derivation (std) of the average result of all clients across the three trials. Besides, we also performed statistical significance testing by conducting paired t-test between each method and our method, and we reported the corresponding p-value. 

Tables~\ref{tab:stab1} and \ref{tab:stab2} shows the results on three datasets, indicating FedRDN achieves a higher mean while maintaining a low std on three datasets compared to other methods. This demonstrates that the randomness introduced by Eq.~\eqref{eq:eq7} in FedRDN does not lead to performance instability. By statistical significance testing, we observed that the p-values across three datasets are less than 0.05, indicating that the performance improvement of FedRDN over other methods is significant. 

% \begin{table}[ht]
% \centering
% \caption{\textbf{Results of Stability} on  Office-Clatch-10~\cite{gong2012geodesic}, DomainNet~\cite{peng2019moment}, and ProstateMRI~\cite{liu2020ms}. We report the mean and standard derivation of the average result of all clients across the three trials: (mean$\pm$std). Besides, we perform paired t-test between each method and our method, and reported the corresponding p-value.}
% \footnotesize
% \setlength\tabcolsep{2pt}
% \renewcommand\arraystretch{1.4}{
% \begin{tabular}{lcccccc}
% \toprule
% \multirow{2}{*}{\textbf{Method}}&\multicolumn{2}{c}{\textbf{Office-Caltech-10}} &\multicolumn{2}{c}{\textbf{DomainNet}}&\multicolumn{2}{c}{\textbf{ProstateMRI}}\\
% \cmidrule(lr){2-3}
% \cmidrule(lr){4-5}
% \cmidrule(lr){6-7}
% & Accuracy & p-value & Accuracy & p-value & Accuracy & p-value \\

% \hline
% \hline
% FedAvg  & 61.74$\pm$0.76 & 0.0001   & 42.66$\pm$1.14 & 0.0128  & 90.44$\pm$0.99 & 0.0237 \\
% \quad+ \textit{norm} & 61.55$\pm$0.40 & 0.0031 & 42.64$\pm$1.55 & 0.0160 &90.60$\pm$0.94 & 0.0035\\
% \quad+ \textit{FedMix} & 63.22$\pm$1.08  & 0.0028& 43.05$\pm$1.43& 0.0047 & - & - \\

%  \rowcolor[HTML]{EFEFEF}  \quad+ \textbf{\textit{FedRDN}} & $\textbf{69.18$\pm$0.63}$ & -&$\textbf{43.94$\pm$1.35}$ & -& \textbf{92.19$\pm$0.86} &- \\

% \bottomrule
% \end{tabular}}
% \label{tab:stab}
% \end{table}

\begin{table}[ht]
\centering
\caption{\textbf{Results of Stability} on  Office-Clatch-10~\cite{gong2012geodesic} and DomainNet~\cite{peng2019moment}. We report the mean and standard derivation of the average result of all clients across the three trials: (mean$\pm$std). Besides, we perform paired t-test between each method and our method, and reported the corresponding p-value.}
\footnotesize
\setlength\tabcolsep{6pt}
\renewcommand\arraystretch{1.4}{
\begin{tabular}{lcccc}
\toprule
\multirow{2}{*}{\textbf{Method}}&\multicolumn{2}{c}{\textbf{Office-Caltech-10}} &\multicolumn{2}{c}{\textbf{DomainNet}}\\
\cmidrule(lr){2-3}
\cmidrule(lr){4-5}
& Accuracy & p-value & Accuracy & p-value \\

\hline
\hline
FedAvg  & 61.74$\pm$0.76 & 0.0001   & 42.66$\pm$1.14 & 0.0128  \\
\quad+ \textit{norm} & 61.55$\pm$0.40 & 0.0031 & 42.64$\pm$1.55 & 0.0160 \\
\quad+ \textit{FedMix} & 63.22$\pm$1.08  & 0.0028& 43.05$\pm$1.43& 0.0047 \\

 \rowcolor[HTML]{EFEFEF}  \quad+ \textbf{\textit{FedRDN}} & $\textbf{69.18$\pm$0.63}$ & -&$\textbf{43.94$\pm$1.35}$ & -\\

\bottomrule
\end{tabular}}
\label{tab:stab1}
\end{table}

\begin{table}[ht]
\centering
\caption{\textbf{Results of Stability} on ProstateMRI~\cite{liu2020ms}. We report the mean and standard derivation of the average result of all clients across the three trials: (mean$\pm$std). Besides, we perform paired t-test between each method and our method, and reported the corresponding p-value.}
\footnotesize
\setlength\tabcolsep{6pt}
\renewcommand\arraystretch{1.4}{
\begin{tabular}{lcc}
\toprule
\multirow{2}{*}{\textbf{Method}}&{\textbf{ProstateMRI}}\\
\cmidrule(lr){2-3}
& Accuracy & p-value \\

\hline
\hline
FedAvg   & 90.44$\pm$0.99 & 0.0237 \\
\quad+ \textit{norm}  &90.60$\pm$0.94 & 0.0035\\

 \rowcolor[HTML]{EFEFEF}  \quad+ \textbf{\textit{FedRDN}} &  \textbf{92.19$\pm$0.86} &- \\

\bottomrule
\end{tabular}}
\label{tab:stab2}
\end{table}

\end{document}